\documentclass[a4paper, 10 pt, conference]{ieeeconf}
\usepackage{comment}
\usepackage{color}
\usepackage{amsmath, amssymb}
\usepackage{graphicx}
\usepackage{etex}
\usepackage{placeins}
\usepackage{afterpage}

\IEEEoverridecommandlockouts  
\usepackage{comment}
\usepackage{siunitx}
\usepackage{relsize}
\usepackage{ifthen}
\usepackage[colorinlistoftodos]{todonotes}

\usepackage[caption=false]{subfig}

\usepackage[vlined,ruled,linesnumbered]{algorithm2e}
\usepackage{graphics} 
\usepackage{rotating}
\usepackage{color}
\usepackage{enumerate}
\usepackage[T1]{fontenc}
\usepackage{psfrag}
\usepackage{epsfig} 
\usepackage{booktabs}
\usepackage{graphicx,url}
\usepackage{multirow}
\usepackage{array}
\usepackage{latexsym}
\usepackage{amsfonts}
\usepackage{amsmath}
\usepackage{amssymb}
\usepackage{xstring}
\usepackage[noend]{algorithmic}
\usepackage{multirow}
\usepackage{xcolor}
\usepackage{prettyref}
\usepackage{flexisym}
\usepackage{bigdelim}
\usepackage{breqn} 
\usepackage{listings}
\let\labelindent\relax
\usepackage{enumitem}
\usepackage{xspace}
\usepackage{bm}
\graphicspath{{./figures/}}
\usepackage{tikz}
\usetikzlibrary{matrix,calc}

\usepackage{mdwlist}

\let\itemize\undefined
\makecompactlist{itemize}{stditemize}

\newrefformat{prob}{Problem\,\ref{#1}}
\newrefformat{def}{Definition\,\ref{#1}}
\newrefformat{sec}{Section\,\ref{#1}}
\newrefformat{sub}{Section\,\ref{#1}}
\newrefformat{prop}{Proposition\,\ref{#1}}
\newrefformat{app}{Appendix\,\ref{#1}}
\newrefformat{alg}{Algorithm\,\ref{#1}}
\newrefformat{cor}{Corollary\,\ref{#1}}
\newrefformat{thm}{Theorem\,\ref{#1}}
\newrefformat{lem}{Lemma\,\ref{#1}}
\newrefformat{fig}{Fig.\,\ref{#1}}
\newrefformat{tab}{Table\,\ref{#1}}

\newcommand{\bdmath}{\begin{dmath}}
\newcommand{\edmath}{\end{dmath}}
\newcommand{\beq}{\begin{equation}}
\newcommand{\eeq}{\end{equation}}
\newcommand{\bdm}{\begin{displaymath}}
\newcommand{\edm}{\end{displaymath}}
\newcommand{\bea}{\begin{eqnarray}}
\newcommand{\eea}{\end{eqnarray}}
\newcommand{\beal}{\beq \begin{array}{ll}}
\newcommand{\eeal}{\end{array} \eeq}
\newcommand{\beas}{\begin{eqnarray*}}
\newcommand{\eeas}{\end{eqnarray*}}
\newcommand{\ba}{\begin{array}}
\newcommand{\ea}{\end{array}}
\newcommand{\bit}{\begin{itemize}}
\newcommand{\eit}{\end{itemize}}
\newcommand{\ben}{\begin{enumerate}}
\newcommand{\een}{\end{enumerate}}

\newcommand{\setal}{~\emph{et~al.}\xspace}

\newcommand{\LC}[1]{{\color{red} \textbf{LC}: #1}}

\newcommand{\hide}[1]{}

\newcommand{\grayout}[1]{{\color{gray} #1}}

\newcommand{\hiddenText}{{\color{gray} hidden text.}}
\newcommand{\hideWithText}[1]{\hiddenText}

\newcommand{\scenario}[1]{{\smaller \sf#1}\xspace}

\newcommand{\blue}[1]{{\color{blue}#1}}

\newcommand{\linkToPdf}[1]{\href{#1}{\blue{(pdf)}}}
\newcommand{\linkToPpt}[1]{\href{#1}{\blue{(ppt)}}}
\newcommand{\linkToCode}[1]{\href{#1}{\blue{(code)}}}
\newcommand{\linkToWeb}[1]{\href{#1}{\blue{(web)}}}
\newcommand{\linkToVideo}[1]{\href{#1}{\blue{(video)}}}
\newcommand{\linkToMedia}[1]{\href{#1}{\blue{(media)}}}
\newcommand{\award}[1]{\xspace} 

\newcommand{\NName}{{\tt ATOM}\xspace}
\newcommand{\MeshName}{\scenario{GraphCMR}}
\newcommand{\featureName}{{\tt FeatureM}\xspace}
\newcommand{\cannyName}{{\tt CannyM}\xspace}
\newcommand{\timeName}{{\tt TimeM}\xspace}
\newcommand{\mrcnn}{{\tt MRcnnM}\xspace}

\newcommand{\PrunePerf}{12.5\%\xspace}
\newcommand{\PruneMinPerf}{126.5\%\xspace}
\newcommand{\Optional}[1]{}
\newcommand{\strike}[1]{}
\newcommand{\AG}[1]{#1}

\newcommand{\eg}{\emph{e.g.,}\xspace}
\newcommand{\ie}{\emph{i.e.,}\xspace}

\newcommand{\Addr}[1]{}

\newcommand{\RGBimage}{RGB image\xspace}
\newcommand{\BinaryMask}{binary mask\xspace}

\newcommand{\noAugmentation}{\scenario{noAug}}

\newcommand{\noJoints}{\scenario{noJoint}}
\newcommand{\noMask}{\scenario{noMask}}

\graphicspath{{../}}

\renewcommand{\NName}{\scenario{ATOM}}

\usepackage{float}
\usepackage[bookmarks=true]{hyperref}
\usepackage{epstopdf}
\usepackage{epsfig}
\usepackage{soul}

\usepackage{xr}

\title{\huge{Online Monitoring for Neural Network Based Monocular Pedestrian Pose Estimation}}

\author{Arjun Gupta and Luca Carlone
\thanks{A.\,Gupta and L.\,Carlone are with the Laboratory for 
Information \& Decision Systems, Massachusetts Institute of Technology, Cambridge, MA, USA, 
{\sf \{argupta,lcarlone\}@mit.edu}
}}

\begin{document}

\maketitle
\begin{tikzpicture}[overlay, remember picture]
\path (current page.north east) ++(-6.7,-1.0) node[below left] {
	\large Accepted for publication at ITSC 2020
};
\end{tikzpicture}

\begin{abstract}
Several autonomy pipelines now have core components that rely on deep learning approaches.
While these approaches work well in nominal conditions, they tend to have unexpected and severe failure modes
that create concerns when used in safety-critical applications, including self-driving cars.
There are several works that aim to characterize the robustness of networks offline, but 
currently there is a lack of tools to monitor
the correctness of network outputs \emph{online} during operation. 
We investigate the problem of online output monitoring 
for neural networks that estimate
 3D human shapes and poses from images.
  Our first contribution is to present and evaluate model-based and learning-based monitors for a human-pose-and-shape 
  reconstruction network, and 
  assess their ability to predict the output loss for a given test input. 
As a second contribution, we introduce an \emph{Adversarially-Trained Online Monitor} (\NName) that learns how to effectively predict losses from data. 
\NName dominates model-based baselines and can detect bad outputs, 
 leading to substantial improvements {in human pose output} quality.
 Our final contribution is an extensive experimental evaluation that shows that discarding outputs flagged as incorrect by \NName 
 improves the average error by
\PrunePerf, and the worst-case error by \PruneMinPerf.
\end{abstract}

\section{Introduction}
\label{sec:intro}

Neural Networks are growing in popularity due to their ability to find meaningful patterns in high-dimensional data such as camera images. 
This ability has made them a primary approach in a variety of tasks ranging from natural language processing 
to vision~\cite{GarciaGarcia17arxiv,He16cvpr-ResNet,Devlin18arxiv-bert, Zou19arxiv-survey} \Addr{add more refs}.  
While these approaches work well in nominal conditions, they  have unexpected and severe failure modes, \eg when 
the distribution of the test data significantly differs from training \cite{Hendrycks16arxiv, liang17arxiv}\Addr{add refs}. 
While many applications can afford occasional failures (e.g., AR/VR, domestic robotics), 
incorrect predictions may have 
disastrous
consequences in 
safety-critical applications: in
May 2017, a Tesla Model S was involved
in a fatal accident in which the autopilot
system did not detect a tractor-trailer~\cite{teslaAccident,teslaAccident2}; in March 2018, a self-driving Uber
vehicle failed to detect a pedestrian and provide a timely warning to the emergency backup driver, killing a woman in
Arizona~\cite{uberAccident}. 

\begin{figure}[]
    \centering
    \includegraphics[width=0.48\textwidth]{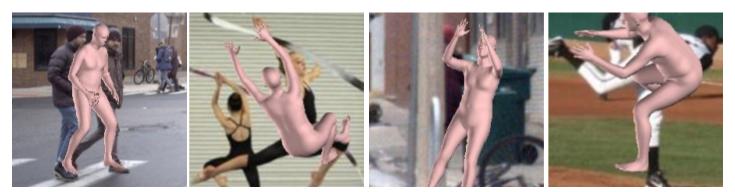}\vspace{-3mm}
    \caption{3D human shape and pose estimation networks are prone to making mistakes when tested outside the training distribution. We discuss how to detect failures and propose an \emph{Adversarially-Trained Online Monitor} (\NName) to detect incorrect outputs. Human pose estimation is crucial for safety-critical applications, including pedestrian detection. 
    \label{fig:example_prune}\vspace{-4mm}}
\end{figure}

While neural networks are powerful, there are no good ways to bound their error for an unseen datapoint, especially one unlike those from training or validation. 
To mitigate this problem, many recent papers attempt to verify the correctness or stability of neural network outputs with respect to specific inputs. A variety of methods 
have already been developed for offline verification \cite{Xiang18arxiv-NeuralVerificationSurvey, Leofante18AutomatedVerification, Xiang17corr-reachableSetComputation, LomuscioM17}; 
however, these methods are limited in the size of the network, type of network, or only find adversarial examples without explicitly bounding the performance in all cases.
These limitations make the existing methods hardly useful for modern day, large-scale machine learning which uses large, complex networks. 
A new set of works attempts to monitor network outputs at runtime to quantify their reliability.
Among these, the works \cite{Balakrishnan19-TQTL,Kang2018ModelAF} rely on hand-crafted constraints and time-series input data for monitoring object detection pipelines.\Addr{what are they monitoring?}
Other works
explore learning-based approaches for  neural network monitoring. 
These approaches range from learning how to predict the correctness of segmentation masks for medical imaging \cite{DeVries18} to false negative detection for road-sign detection 
\cite{RahmanIROS19-falseNegative}. 

\begin{figure*}[h!]
	\centering
	\includegraphics[width=1\textwidth]{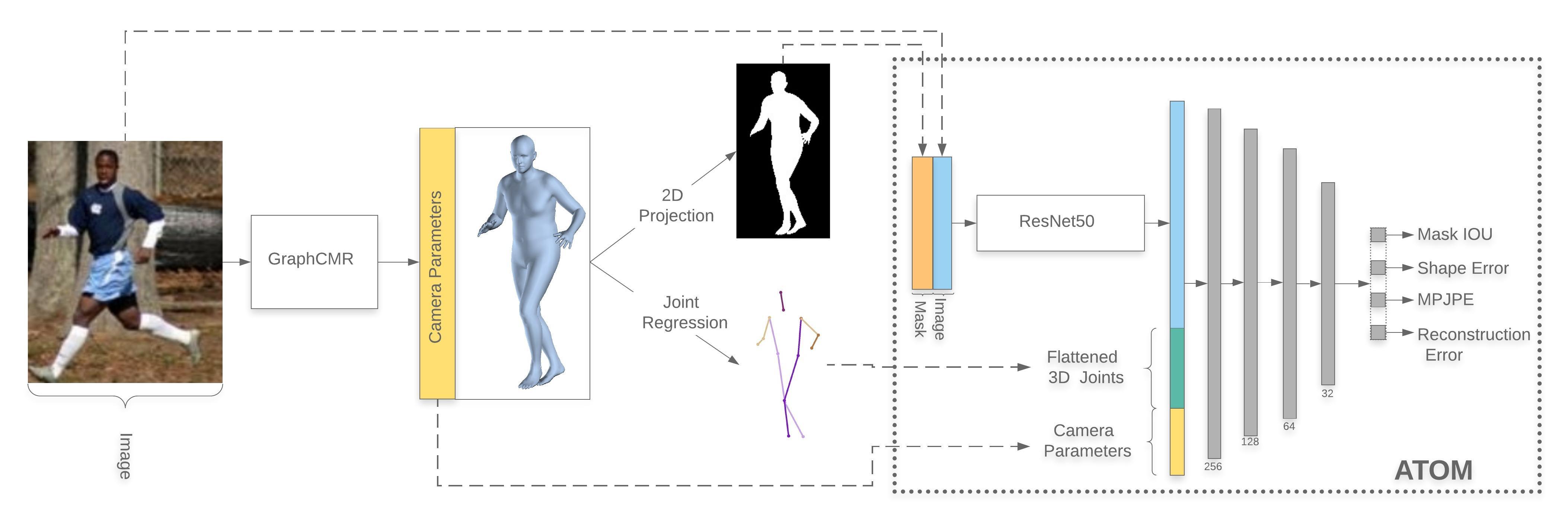}\vspace{-3mm}
	\caption{We propose \NName, a learning-based approach to monitor the outputs of a 3D human shape and pose estimation network 
    (\MeshName~\cite{Kolotouros19cvpr-shapeRec}). At test time, \NName takes as input the original input image, the 2D projection
     of the 3D mesh produced by \MeshName, camera parameters output by \MeshName, and 3D joints extracted from the 3D mesh. 
     It then predicts the quality (\ie the loss) of the provided mesh model.
	\label{fig:method-depiction}\vspace{-4mm}}
\end{figure*}

In this paper, we propose the first approach to monitor the quality of 3D human pose estimates from a neural network. Human pose estimation is crucial for pedestrian detection in intelligent transportation, as well as in safety-critical applications 
of robotics (\eg collaborative manufacturing).
We develop several model-based monitors to detect incorrect human pose predictions.
We also employ a pre-trained 2D human segmentation network to monitor the 3D network outputs.
As a second contribution, we develop an \emph{Adversarially-Trained Online Monitor} (\NName), which is trained end-to-end to predict losses on the 3D human (shape and) pose  outputs. 
A depiction of the overall structure of \NName for 3D human pose monitoring is shown in Fig.~\ref{fig:method-depiction}. 
Finally, we provide an extensive experimental evaluation across four benchmarking datasets. Our results show that, due to the complexity of the problem, model-based  approaches have limited success at monitoring the network outputs. 
Learning-based approaches provide a significant improvement
over model-based \mbox{methods and \NName dominates the compared techniques}. 

In summary, we propose three key contributions: 
\begin{enumerate}
	\item We develop and implement several baseline methods for loss prediction and online monitoring of 3D human pose estimates. These baselines include both model-based and learning-based methods leveraging existing networks;
    \item We develop a new architecture, named \emph{Adversarially-Trained Online Monitor} (\NName),  to directly predict different losses of the estimated 3D human pose outputs.
    \item  We present an extensive experimental evaluation that shows that learning-based 
    monitors are extremely effective in predicting the output loss and at detecting bad network predictions.
\end{enumerate}

In the following, we review related work (Section~\ref{sec:related_work}) and present several baseline techniques for output monitoring  (Section~\ref{sec:baselines}). 
 Then, we introduce \NName as a more effective learning-based monitor (Section~\ref{sec:NName}). 
 Finally, we present the results of our experimental analysis (Section~\ref{sec:experiments}) and draw conclusions (Section~\ref{sec:conclusion}).


\section{Related Work}
\label{sec:related_work}
We review related work dealing with network verification and monitoring and refer the reader to~\cite{Xiang18arxiv-NeuralVerificationSurvey, Leofante18AutomatedVerification} 
for a more extensive survey. 
 Seshia\setal~\cite{Seshia16arxiv} provide\strike{s} an excellent overview of why verifying neural networks is 
  much more difficult than verifying other types of programs and outline the technical advances necessary to verify them.

\subsection{Offline Verification}

 \subsubsection{Reachability Analysis} 
  A common approach for neural network verification is to analyze the \emph{reachable set}, \ie the set of possible network outputs for a given set of inputs, to assess when the network outputs unsafe values \cite{Xiang17corr-reachableSetComputation,
    LomuscioM17, Xiang17corr}. Katz\setal~\cite{Katz17coRR} uses an SMT solver to ensure that under certain preconditions, the outputs of the neural network obey a set of constraints.
    \Addr{add work from Ziko Kolter and George Pappas} Fazlyab\setal~\cite{Fazlyab19arxiv-verificationNN} abstract activation functions in the form of quadratic constraints and solve the resulting semidefinite program to find the 
    reachable output set of a network.
    Contrarily to offline verification, our goal is to verify the network outputs during operation.

\subsubsection{Adversarial Perturbations}
Another major approach for quantifying the robustness of a network is to find the minimum \emph{adversarial perturbation} that causes the network to make significant errors. Bastani\setal~\cite{Bastani16},
   Cheng\setal~\cite{ChengNR17}, and Sun\setal~\cite{Sun18} attempt to find the adversarial perturbation with minimum norm that causes a network to misclassify examples. Huang\setal~\cite{HuangKWW16} guarantee that they find an adversarial example (if it exists)
   for a network to a particular class of pertubation. AI2~\cite{Gehr18ieee-Ai2} attempts to solve problems of 
   scalability and precision for perturbation analysis by using the theory of abstract interpretation to capture all variations of a perturbation at once. Kolter\setal~\cite{Kolter17arxiv-provableDefense} learn a classifier that is robust to 
   any norm-bounded adversarial attack. DeepXplore~\cite{Pei17-DeepXplore}, and DeepGauge~\cite{Ma18-DeepGauge} find adversarial examples by perturbing images to maximize neuron coverage and cause the network to
   misclassify examples. Sun\setal~\cite{Sun18-concolic_testing} discover adversarial perturbations using a concolic testing paradigm that alternates between concrete and 
   symbolic execution to generate adversarial examples. Dvijotham\setal~\cite{Dvijotham18-dual_approach_verification} formulate adversarial
   perturbations as an optimization problem and approximate the dual to get an upper bound on the adversarial error rate. Wong\setal~\cite{Wong18neurips-provableDefense} attempt to scale the process of making networks robust to adversarial examples to 
   deep learning architectures.

    \subsubsection{Test Case Generation}
 The above methods all test the network by finding some simple perturbation (\eg additive pixel-wise noise) of an existing image that the network fails to identify correctly. 
 However, this is limited in scope, and it doesn't guarantee correctness on images that have large, structural dissimilarity to the training data. 
 Some recent papers have identified other solutions to verifying networks against more structured and realistic perturbations.
  DeepTest~\cite{Tian17-DeepTest} automatically modifies images for self-driving in a structured way (\eg by adding rain, fog) and searches for input images that maximize the network neuron 
  coverage. 
  Fremont\setal~\cite{Fremont18-SCENIC}, and Cumhur\setal~\cite{Cumhur18} develop frameworks to specify arbitrary, randomized scenarios for autonomous driving and generate photo-realistic images for 
  testing and training networks. 
  While these pipelines are able to
  synthesize completely new images, they both require the user to manually specify test cases, implicitly
   reducing their ability to find bad samples.

\subsection{Online Monitoring}


\subsubsection{Temporal Logic}
 Balakrishnan\setal~\cite{Balakrishnan19-TQTL} use Timed Quality Temporal Logic (TQTL) to reason about logical relations between boolean-value propositions on the network output over time. 
 Kang\setal~\cite{Kang2018ModelAF} present the idea of model assertions, which similarly place a constraint on the model outputs to detect when they violate certain condition\AG{s}. 
 These approaches add constraints on the network outputs, but require hand-made boolean constraints. Moreover, they require the network to predict 
 over several timesteps before providing a feedback. 

\subsubsection{Learning-based Monitoring}
Learning-based monitoring is most closely related to our proposal. 
DeVries\setal~\cite{DeVries18} develop a method for verifying a semantic segmentation network for medical imaging by passing the original image, the segmentation, 
and a confidence mask to an external verification network. Hecker\setal~\cite{Hecker18} create a pipeline to predict whether an autonomous driving system will make bad 
predictions in the future. 
\Addr{keep style "LastName\setal~[x] bla bla"}
Saxena\setal~\cite{Saxena17} and Daftry\setal~\cite{DaftryZBH16} develop a network to verify
 collision-free trajectories for a vision-based drone platform. Rahman\setal~\cite{RahmanIROS19-falseNegative} process the hidden layer outputs of a neural 
 network to predict when a traffic sign detection network outputs false negatives. 
 Cheng\setal~\cite{Cheng18date} and Henzinger\setal~\cite{Henzinger19arxiv} observe neuron activation patterns to monitor when the network is operating on inputs unlike the data seen during training.
  Hendrycks\setal~\cite{Hendrycks16arxiv} develop a method of monitoring network confidence based on softmax probabilities. Our model works on networks without softmax activations, and 
 gives finer-grained monitoring information (\ie we predict the future loss for a given input). 
  In addition, none of these works tackle the problem of monitoring 3D human pose estimates, which is key to autonomous vehicles.

\section{Online Monitoring of 3D Human Poses: Problem Statement and Baseline Approaches}
\label{sec:baselines}


\subsection{Problem Statement}

We address the problem of monitoring \MeshName~\cite{Kolotouros19cvpr-shapeRec}, a state-of-the-art human shape and pose reconstruction
network. \MeshName takes images as input and outputs the SMPL mesh~\cite{Loper15tg-smpl} of the center-most human. More practically, we solve the problem of detecting when the SMPL mesh output 
from \MeshName does not align with the input image. Our approach is to design a monitor that provides a score of mesh correctness
based on how it compares to the input image. We then use those scores to detect when a mesh might be incorrect. Before we proceed with our contributions, we first define the
 language used throughout the section:

\subsubsection{Skinned Multi-Person Linear Model (SMPL)}
We make frequent references to SMPL~\cite{Loper15tg-smpl}, the output of \MeshName. SMPL parametrizes a full human mesh using a set of pose parameters, $\Theta$, which
define the relative angle-axis rotation between joints and a set of body-shape parameters $\beta$, which describe the way the human appears. From an SMPL model, we can also regress 
the 3D human joint-locations, which we use in our monitoring approaches.

\subsubsection{Human Shape and Pose Estimation} 
\MeshName is an example of a human-shape-and-pose estimation approach, which fits an SMPL mesh to a monocular image. Previous human shape and pose estimation techniques include
SMPLfy \cite{Bogo16eccv-keepItSMPL}, HMR~\cite{Kanazawa18cvpr-humanPoseAndShape}, NBF~\cite{Omran183dv-humanPoseAndShape}, as well as many others
 \cite{Zanfir18cvpr-humanPoseAndShape, Ramakrishna12eccv-humanPose, Wang14cvpr-humanPose, Cho17cviu-humanPose, Zhou14eccv-humanPose}. 
   We design a monitor for \MeshName because it provides state-of-the-art performance, 
   although it still fails on more difficult input instances. 

We proceed by describing our baseline approaches for human pose monitoring, including approaches that use
existing neural network models.

\subsection{Feature Matching (\featureName)}
\begin{figure}[h!]
    \centering
    \includegraphics[width=0.47\textwidth]{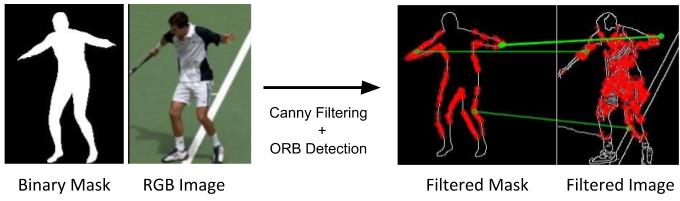} \vspace{-4mm}

    \caption{\featureName performs edge detection on a given \RGBimage and on the \BinaryMask obtained by projecting the 3D mesh by \MeshName.
     It then compares features in the edge images to assess the mesh quality. 
        The red circles are all detected features, and the 
        green circles are the features that are considered inlier matches.}
    \label{fig:feature_matching}
\end{figure}

Our first approach projects the estimated 3D mesh into a 2D \BinaryMask using Neural 3D Mesh Renderer \cite{Kato18cvpr} and compares it to the original image using feature extraction and matching (Fig.~\ref{fig:feature_matching}). 
Classically, SIFT~\cite{Lowe04ijcv}, SURF~\cite{Bay06eccv}, and ORB~\cite{Rublee11iccv-ORB} are commonly used for object detection in cluttered images.
We adapt ORB to detect features for our monitoring application.
While {ORB} features work well when comparing two RGB images, they {fail when} comparing an \RGBimage with a \BinaryMask.
To resolve this issue, 
we first extract contours using the Canny detector~\cite{Canny86pami} for both the 
\RGBimage and the binary mask and then perform feature detection and matching on the contours.

We match keypoints from the \BinaryMask to the two nearest keypoints in the \RGBimage based on feature descriptors. 
Because the mask and the original image should have the human at the same location, we additionally use the pixel
distance between keypoints to determine inlier matches. 
We count a match as an inlier if the descriptor distance to the \AG{matched keypoint is less than $0.75$ times the descriptor distance to the next closest keypoint
 and the keypoints are within $26$ pixels of each other (determined experimentally)}.
We score the image correctness based on the percentage of keypoints in the 2D \BinaryMask that correspond to features in the original \RGBimage. Fig.~\ref{fig:feature_matching} shows 
that this approach, \AG{that we denote as \featureName, picks out joints along the silhouette} (the most feature-rich regions) and uses the corresponding mismatch to compute an error score.

\Optional{While reasonable in theory, this approach has several failure modes in the cases where the background is cluttered and the edge detection picks up a lot of noise.
In addition, when poses are even minorly different (and thus change feature descriptors significantly), this method is unable to provide much information on the 
quality of the prediction.}

\subsection{Canny Edge Matching (\cannyName)}
Our second monitoring baseline is based on the Canny edge detector.
The approach, named \cannyName, is motivated by the observation that the 2D projection of the human mesh should precisely outline the human silhouette in the image.
We once again apply Canny filtering to both the \BinaryMask
 and the original \RGBimage to get contours. We then dilate the contours from the original input image using a fixed size kernel 
(a $5\times5$ kernel in our tests) and measure the percentage of the contour points from the 2D human \BinaryMask that overlap with a dilated contour from the \RGBimage. The percentage overlap is used as score by the monitoring approach.

\Addr{the following paragraph seems to belong to the experimental evaluation}
\Optional{This method suffers when images are low-contrast with the human (resulting in poor edges) or the background is cluttered.} 


\subsection{Shape and Pose Consistency (\timeName)} 
For timeseries data, a clear baseline is to judge the consistency of predictions over time. These consistency metrics are similar in spirit to TQTL~\cite{Balakrishnan19-TQTL} where
we reason about the consistency of the relationship between consecutive pose predictions. 
 We assume that 
 \AG{we} can correctly
 \AG{re-identify the same human across multiple images:}
the datasets we use for evaluation provide an ID for each annotated human.\footnote{In practice, data association would be based on temporal tracking~\cite{Cadena16tro-SLAMsurvey}.} 
This third baseline approach, named \timeName, is applied independently to each ID, and 
leverages the loss functions proposed by Kim\setal~\cite{Kim18arxiv-PedX}.

\AG{The \timeName score} is the sum of two terms: \emph{shape consistency} and \emph{pose consistency}. \emph{Shape consistency} enforces the observation that the body shape of
a person is constant over time. \emph{Pose consistency} ensures that the difference between consecutive poses corresponds to a feasible human motion. We formalize shape and pose consistency below.

\paragraph{Shape Consistency} For shape consistency, we maintain a database of the average body shape seen for each person ID. Each time
we observe the same ID, we compute the difference between the predicted body-shape parameters and the stored average for that ID to get a similarity score. We 
update the average for future use by averaging the new detection body shape with the current average. Shape Consistency evaluates the correctness of a mesh based on its deviation from the average body shape for that subject.
The shape consistency metric is defined as:

\begin{equation}
\label{eq:shapeScore}
    \mathcal{L}_{sc} = ||\beta_{t} - \beta_{avg}||_F^2
\end{equation}
Where $\beta_{t}$ is a matrix of the body shape parameters for the detection at time $t$ and the $\beta_{avg}$ is a matrix of the average body shape for that ID. The symbol $||\cdot||_F$ denotes the Frobenius norm.

\paragraph{Pose Consistency} Unlike shape consistency, pose consistency is measured between adjacent predictions only and captures the change in pose due to joint movements. 
To measure pose consistency, we measure the distance between corresponding 3D joints in the current and previous timestep. We weight each joint error
based on its distance from the core body (e.g. an elbow cannot move as quickly as a hand) to reduce the larger errors from more distant appendage points. \Addr{give expression of W} We additionally include a history buffer which
compares the current pose against previous poses from a number of timesteps back in the case that the previous errors were high. This ensures that a bad prediction in a sequence of poses does not cause future good 
predictions to have high error. We formalize this below:

\begin{equation}
\label{eq:poseScore}
    \mathcal{L}_{pc} = \min_{0 \leq i \leq h}(J_{t-i} - J_t)^T W (J_{t-i} - J_t)
\end{equation}

Where $h$ is the history length, $J_t$ are the joints observed at time $t$, and \AG{$W$ is a diagonal matrix of joint weights.}
The loss can be thought of as enforcing a prior that humans tend to have very small changes between consecutive poses.

\timeName scores each output using the overall score $\mathcal{L}_{sc} + \mathcal{L}_{pc}$.
Since we do not know all the detections of each person ID beforehand, the first time we encounter a new ID, we give them a $0$ score
since we have no prior experience to use to compute the scores~\eqref{eq:shapeScore}-\eqref{eq:poseScore}. As we observe more instances of the ID, we increase our ability to discern good predictions from bad.

\subsection{MaskRCNN Intersection Over Union (\mrcnn)} \label{sec:iou}
The above approaches use purely classical model-based methods to monitor the 3D human meshes produced by \MeshName; as we will see, these techniques perform poorly in practice. 
The first two
approaches (\featureName and \cannyName) tend to fail on images with cluttered backgrounds, where
edge detection returns contours where the human is difficult to distinguish. 
\timeName, on the other hand, does not apply when there is no prediction history.
This section solves these issues by leveraging a learning-based 2D pixel-wise segmentation network (we use MaskRCNN~\cite{He17iccv-maskRCNN})
 to detect failures in 3D human mesh outputs by \MeshName.

This monitoring approach, named \mrcnn, uses the predicted mask from MaskRCNN as a pseudo ground-truth to verify the correctness of \BinaryMask obtained from \MeshName. We compute the score for a \MeshName output as the Intersection over
Union of the MaskRCNN mask and the \BinaryMask obtained by projecting the 3D mesh:

\begin{equation}
    \mathcal{L}_{IOU} = \frac{2}{|M_{m}| + |M_{p}|} \sum_{(i, j) \in M_p} \mathbb{I}\{M_p(i, j) = M_{m}(i, j)\}
\end{equation}

Where $M_{m}$ is the mask produced by MaskRCNN,  $M_{p}$ is the predicted mask, $(i, j)$ are pixel coordinates, and 
$\mathbb{I}$ is the indicator function which is 1 when
 $M_p(i, j) = M_{m}(i, j)$ 
 and zero otherwise. Because there could be potentially many humans, and the mesh prediction network only predicts one, we choose
the highest IOU score. 

At a first glance, using a neural network to monitor the correctness of another network seems circular, as the monitor network itself is bound to be unreliable. However, the results show that this is in fact a much more effective
method than the classical approaches. 
The intuition is that training MaskRCNN is much easier than training \MeshName, since the former only requires labeling a 
2D image (relatively easy for a human), while the latter requires training on 3D meshes, which are difficult to label at scale. Therefore, MaskRCNN has access to a much larger training set which allows it to perform well
 on instances that cause failures in \MeshName. 

While the \mrcnn approach can successfully solve the problems faced by the model-based approaches, it is still unreliable when monitoring the worst samples (as seen in Section \ref{sec:results}).
In addition, \mrcnn is unable to accurately predict losses on the SMPL pose, a critical metric for correctness.
This motivates us to design a novel approach, \NName, presented below.


\section{\NName: Adversarially-Trained Online Monitor}
\label{sec:NName}

While the previous monitors made use of model-based approaches or existing neural networks, 
\NName is taught end-to-end to predict losses on \MeshName outputs.
Our network takes as input:
(i) a greyscale version of the original \RGBimage concatenated with the \BinaryMask of the predicted mesh,
(ii) the 3D joint locations, and 
(iii) the camera parameters output by \MeshName.
\NName simultaneously outputs the predicted 
\emph{MaskIOU} between the
\BinaryMask and a theoretical ground-truth mask, 
the \emph{Mean Per Joint Prediction Error} (MPJPE)~\cite{Zhou18PAMI-monocap},
the \emph{Reconstruction Error} (REC)~\cite{Zhou18PAMI-monocap}, 
and \AG{\emph{Shape Error} which is the \mbox{per-vertex error of the output mesh.}}

\subsubsection{Network Architecture}
We depict our model architecture in Fig.~\ref{fig:method-depiction}. We concatenate the greyscale image and the \BinaryMask along the channel axis and use a Resnet50 architecture to encode their combination. We remove the 
last layer and use the final encoded vector through the rest of the network. We separately flatten the 3D joint locations and concatenate them with the 
encoded image and the camera parameters. We pass the concatenated vector through a set of fully-connected layers \AG{to regress the loss on \MeshName outputs}.
 Notice that our model architecture is relatively shallow for the deep learning era. This allows it to run at 10 Hz on a modest GPU, meaning that it can reasonably be run alongside another
network and give real time feedback at low computational cost. 

\subsubsection{Loss Functions}
For training loss, we use the Mean Squared Error (MSE) between the network output and the ground-truth:

\begin{equation}
	\mathcal{L}_{\mathrm{mse}} = \frac{1}{n}\sum_i(\hat{y}_i - y_i)^2
\end{equation}

Where $\hat{y}_i$ is the prediction of \NName for the $i^{th}$ data-point, $y_i$ is the ground-truth loss for that mesh, and $n$ is the number of samples.

\subsubsection{Mesh Augmentation for Learning Approaches} \label{sec:augmentation}
\AG{Because we only monitor the correctness of labels, we can dramatically increase our training data by perturbing \MeshName outputs to get 
new loss labels.}
SMPL meshes can easily by perturbed by perturbing the mesh parameters. This makes data augmentation simple and quite robust. During training,
we randomly perturb a batch of meshes with a predetermined probability ($0.6$ for the model reported in the results) by adding Gaussian noise to the body-shape and
 pose-parameters of the predicted mesh. We observe that this leads to a significant improvement in out-of-sample monitoring.

\newcommand{\PrunePercentile}{80\%}

\section{Experiments}
\label{sec:experiments}

We present a three-part evaluation of the baseline monitors and \NName. Section~\ref{sec:loss_pred}  demonstrates that we can effectively predict output error metrics for \MeshName.
Section~\ref{sec:ablation} justifies our architecture and the effect of data augmentation via an ablation study.
Finally, Section~\ref{sec:results} describes the use of \NName for monitoring \MeshName outputs with quantitative and qualitative results. Before the evaluation, we describe our datasets and evaluation metrics.

\subsection{Datasets and Evaluation Metrics}
The human mesh prediction model~\cite{Kolotouros19cvpr-shapeRec} is evaluated on three datasets: Leeds Sports Poses (LSP) \cite{Johnson10bmvc-LSP}, Unite the People (UP-3D)~\cite{Lassner17cvpr-humanPoseAndShape}, and Human3.6M~\cite{Ionescu14pami-human36m}. Each dataset has its own set of evaluation metrics. 
LSP uses \AG{\emph{Mask Accuracy}} with annotation from the UP dataset, UP-3D uses \AG{\emph{Shape Error}}, and Human3.6M uses \AG{\emph{MPJPE}} and \AG{\emph{REC} (see section \ref{sec:NName})} to evaluate model performance.
We add an additional dataset, PedX \cite{Kim18arxiv-PedX}, for an out-of-distribution comparison, and evaluate the projected 2D masks on this dataset as well. 
We use \AG{\emph{Mask IOU}} (as defined in Section~\ref{sec:iou}) instead of \AG{\emph{Mask Accuracy}}
because it is more consistent across LSP and PedX, and it more accurately captures mask correctness since it does not consider free-space pixels. 

We train \NName exclusively on the test set for \MeshName. We use the first $50\%$ of test data for training, $10\%$ for validation, and evaluate on the remaining $40\%$. We do not train on 
PedX and use all of PedX for evaluation.

\subsection{Loss Prediction} \label{sec:loss_pred}

We first evaluate the monitors discussed in this paper by reporting how their predicted loss (for a given input) 
correlates with the actual loss, computed using the ground truth labels.

\begin{table}[h!] 
	\centering
	\caption{Loss Correlation on LSP, UP-3D, Human3.6M, and PedX}
    \begin{tabular}{cccccc}
        \toprule
         & UP-3d & LSP & \multicolumn{2}{c}{Human3.6M} & PedX\\
		Method & Shape & Mask IOU & MPJPE  & REC & Mask IOU \\
		\toprule

		\featureName &  0.08  & 0.09 & 0.17 & 0.19 & 0.16 \\

		\cannyName & 0.11 & 0.51 & 0.19 & 0.21 & 0.20\\

		\timeName &  n/a & n/a & 0.19 & 0.17 & 0.09 \\
        \mrcnn &  0.29 & 0.75 & 0.28 & 0.21 & \bf 0.80 \\ 
		\NName    & \bf 0.46   & \bf  0.85 &  \bf 0.67 & \bf 0.69 & 0.75 \\
		\bottomrule
	\end{tabular}

    \label{tab:performance}
\end{table}

Table~\ref{tab:performance} shows that the model-based approaches produce scores that are minimally correlated with the ground-truth losses. \featureName has exceptionally low
correlations, likely because of the many failure modes of the feature detector in cluttered images. \cannyName stands out with non-trivial correlation on LSP where
the humans tend to be higher-resolution and better contrasted with the background (allowing cleaner human edge extraction). The performance drops significantly on PedX,
likely due to reduced resolution and contrast in the PedX data and background clutter from the industrial setting. \cannyName clearly beats \featureName on 
all tasks, which is expected because \featureName has the exact same failure modes as \cannyName (when human edges are indistinguishable from background clutter)
with the additional failure mode from bad feature detections.
\timeName performs reasonably when predicting the pose errors, as expected since it computes pose consistency, but it fails to predict the less directly related MaskIOU metric for PedX.
\AG{Moreover, \timeName cannot be applied to UP-3D or LSP because they are not timeseries.}

Table~\ref{tab:performance} also shows that \mrcnn exhibits a 
 substantially higher correlation score. \mrcnn is the first approach, among the ones discussed in this paper, that accurately predicts loss. It even
beats \NName on PedX, although we later show that \NName is much more effective at detecting the worst predictions for PedX. In addition, notice that all baseline 
methods, including \mrcnn, are unable to predict the pose or shape metrics \AG{because they either do not have the information to judge these losses (\mrcnn, \cannyName, \featureName), or 
they can only compare predictions over time without reference to the original image (\timeName).}

On the other hand, \NName is able to generate losses that are very strongly correlated with the ground-truth losses.
The correlation is lower for \AG{\emph{Shape Error}}: this is expected since \emph{Shape Error} 
depends on the location of individual vertices in the mesh, which \NName does not have access to. 
However, the network is able to get high correlations with
the \AG{pose error metrics (\emph{MPJPE} and \emph{REC})} on Human3.6M, a key metric for pedestrian detection monitoring, where the pose is more important that the shape of the human.

\subsection{Ablation Study}\label{sec:ablation}

In this section, we systematically remove key components of \NName and see that each design choice has an immediate effect at improving network correctness.
The model names and their corresponding features are listed below:
\begin{enumerate}
	\item [(i)] \noAugmentation: \NName without augmentation during training. 
	\item [(ii)] \noJoints: \NName without 3D joint inputs.
	\item [(iii)] \noMask: \NName without \BinaryMask inputs.
\end{enumerate}

\begin{table}[h!] 
	\centering
	\caption{Ablation study: loss correlation for \NName}
    \begin{tabular}{cccccc}
        \toprule
         & UP-3d & LSP & \multicolumn{2}{c}{Human3.6M} & PedX\\
		Method & Shape & Mask IOU & MPJPE  & Rec. & Mask IOU\\
		\toprule
		\NName          & \bf 0.46   & \bf  0.85 &  0.67 & \bf 0.69 & \bf 0.75 \\
        \noAugmentation & \bf  0.46  & 0.69      &  0.67  & 0.67     & 0.46 \\
		\noMask         & 0.33       & 0.443  & 0.56      &  0.57    & -0.033 \\
		\noJoints       & 0.44      & 0.791   &\bf 0.69   &  0.67        & 0.606\\
		\bottomrule
	\end{tabular}

    \label{tab:ablation}
\end{table}

Table~\ref{tab:ablation} shows that
the segmentation mask input is integral for predicting any of the metrics successfully as shown by the \noMask performance. 
The augmentation and joints drastically improve the out-of-sample correlation with PedX as well as many of the key in-sample 
performance indicators. 

\begin{table*}[h!] 
	\centering
	\caption{\% Performance increase after pruning outputs with high predicted loss on LSP, UP-3D, Human3.6M, and PedX}
    \begin{tabular}{cccccccccccc}
        \toprule
         & \multicolumn{2}{c}{UP-3D} & \multicolumn{2}{c}{LSP} & \multicolumn{4}{c}{Human3.6M} & \multicolumn{2}{c}{PedX}& \\
		 & \multicolumn{2}{c}{Shape} & \multicolumn{2}{c}{Mask IOU} & \multicolumn{2}{c}{MPJPE} & \multicolumn{2}{c}{Rec.} & \multicolumn{2}{c}{Mask IOU} & Average Prediction Time (s)\\
	    Method	& avg $\downarrow$ & max $\downarrow$ & avg $\uparrow$ & min $\uparrow$ & avg $\downarrow$ & max $\downarrow$ & avg $\downarrow$ & max $\downarrow$ & avg $\uparrow$ & min $\uparrow$ \\ 
		\toprule
        \featureName& 0.0\% & 0.0\% &     0.4\% &  25.6\%&      2.5\% & 0.0\%   & 3.6\%  & 0.0\%     &0.4\% & 1.2\% & 0.012\\ 

        \cannyName & 4.2\% & 0.0\%       & 2.3\%&  0.0\% &      2.4\% & 0.0\%   & 3.6\%  & 0.0\%     & 0.5\% & 0.0\% & 0.43\\


        \timeName &  - & - & - & -                            & 3.7\% & 0.0\%   & 3.6\%  & 0.0\%      & 0.5\% & 0.0\% & 0.0015\\ 

        \mrcnn &  \bf 12.5\% & \bf 21.9\%    & \bf 3.7\% &  25.6\%    & 4.9\% & 53.4\%  & 5.4\%  & \bf 26.1\%     & \bf3.5\% & 0.0\% & 1.25\\ 

        \NName & 10.4\% & \bf 21.9\% & \bf 3.7\% & \bf 87.5\%  & \bf 11.1\% & \bf 70.0\%  & \bf 12.5\%  & \bf 26.1\% & 3.4\% & \bf 126.5\% & 0.0086\\
		\bottomrule
	\end{tabular}

    \label{tab:prune_perf}
\end{table*}

\begin{figure*}[h!]
    \label{fig:prune_examples}
    \centering
    \includegraphics[width=\textwidth]{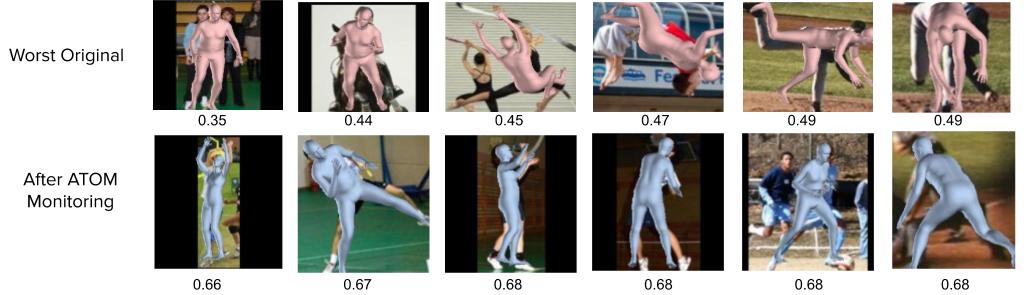}
    \caption{ \AG{We visualize the \MeshName outputs with the worst losses on LSP and the corresponding worst outputs in the set after applying \NName monitoring and removal. The Mask IOU score for each prediction is noted under the image.}\label{fig:prune_examples}}
\end{figure*}

\subsection{Monitoring: Results and Timing} \label{sec:results}

We now present results produced by \NName and baseline methods when monitoring \MeshName. We show the effect of monitoring by removing outputs that are in the worst 20\% of scores (losses) computed by 
our monitoring methods, and we enumerate the percent improvement for each evaluation metric. 
In other words, we evaluate how the error metrics of \MeshName improve when rejecting bad outputs as detected by each monitor.

Table~\ref{tab:prune_perf} shows the percentage improvement in the average error (``avg'' in Table~\ref{tab:prune_perf}) and the worst-case error (``min'' or ``max'', depending on the metric, in Table~\ref{tab:prune_perf})  for each of the discussed monitors.
The table shows that \NName is able to significantly improve the average error metrics by up to \PrunePerf{}. \NName is also able to consistently improve the
error value for the worst sample, improving the minimum MaskIOU on PedX by \PruneMinPerf{}. MaskRCNN does provide some marginally better improvements in PedX and UP-3D; however,
it is extremely inconsistent and cannot reduce the worst values for LSP or PedX.


In addition to having the ability to effectively monitor predictions, \NName's small architecture allows it to be run in real time. The last column in Table~\ref{tab:prune_perf} shows that \NName can run in milliseconds on a 
1060 Geforce Mobile GPU. This observation suggests that \NName can be effectively used for online monitoring since it does not add a significant computational burden.

Fig.~\ref{fig:prune_examples} provides a visualization of the outputs deemed incorrect by \NName.
In particular, the figure shows 
the worst predictions on the LSP test set and the corresponding worst predictions
after rejecting bad outputs using \NName. We provide the MaskIOU score for the prediction under each image.
Notice that the original predictions include several examples where the mesh is completely
misaligned with the actual human and frequently the mesh is not facing the correct direction, an important factor for determining where a human may move next. On the other hand, after monitoring and removal,
all predictions have significantly higher score and match the provided image with a visibly correct body orientation and pose.

\section{Conclusion}
\label{sec:conclusion}

In this paper, we discussed techniques for monitoring a 3D human pose and shape reconstruction network. We demonstrated several model-based approaches, and showed that they 
cannot effectively predict the loss or detect bad outputs. 
We found that networks that predict a simpler form of output (either directly the loss or the 2D segmentation mask) can effectively monitor a network
with more complex, 3D, outputs. We developed a monitoring network, \NName, that dominates the baseline approaches in terms of accuracy and runtime. In future work, we plan to explore the use of Graph Convolutional Networks as an alternative architecture and apply
\NName as a weak supervisor to improve the performance of \MeshName. As systems become more advanced and need to interact with humans, accurate monitoring of human detections and pose estimation becomes a prerequisite for safety-critical applications, and 
\NName provides the first example of online 3D monitoring for these applications. 

\bibliographystyle{IEEEtran}
\bibliography{references/refs,references/myRefs}

\end{document}